\documentclass[letterpaper]{article}
\usepackage{uai2020}
\usepackage[margin=1in]{geometry}

\usepackage{microtype}
\usepackage{graphicx}
\usepackage{subfigure}
\usepackage{booktabs} % for professional tables
\usepackage{adjustbox}
\usepackage{multirow}
\usepackage{caption}
\usepackage{algorithm}
\usepackage{algorithmic}
\usepackage{amsthm}
\usepackage{amssymb}
\usepackage{multirow}
\usepackage[color=green!30]{todonotes}
\usepackage[round,sort,comma,super,authoryear]{natbib}
\usepackage{amsmath,amsfonts,bm}

\def\eqref#1{equation~\ref{#1}}
\def\floor#1{\lfloor #1 \rfloor}
\def\1{\bm{1}}

\def\va{{\mathbf{a}}}
\def\vb{{\mathbf{b}}}

\def\vv{{\mathbf{v}}}

\def\vx{{\mathbf{x}}}

\def\mA{{\bm{A}}}

\def\mF{{\bm{F}}}

\def\mM{{\bm{M}}}

\def\mX{{\bm{X}}}

\DeclareMathAlphabet{\mathsfit}{\encodingdefault}{\sfdefault}{m}{sl}
\SetMathAlphabet{\mathsfit}{bold}{\encodingdefault}{\sfdefault}{bx}{n}

\def\gG{{\mathcal{G}}}

\def\sR{{\mathbb{R}}}

\def\emA{{A}}

\newcommand{\frob}[1]{\left\Vert #1 \right\Vert} % frobenius norm
\newcommand{\tran}[1]{#1^\top}

\newcommand{\pp}{\hspace{4pt} | \hspace{4pt}}

\newcommand{\hmA}{\hat{\mA}}
\newcommand{\hmX}{\hat{\mX}}

\newcommand{\cN}{\mathcal{N}}
\newcommand{\cO}{\mathcal{O}}

\newcommand{\vZero}{\mathbf{0}}

\newcommand{\vdelta}{\mathbf{\delta}}
\newcommand{\amb}[1]{#1^{\prime}}
\newcommand{\ambb}[1]{#1^{\prime\prime}}

\newcommand{\lij}{\ensuremath{\{i,j\}}}

\newcommand{\lkl}{\ensuremath{\{k,l\}}}
\newcommand{\lil}{\ensuremath{\{i,l\}}}

\newcommand{\hV}{\hat{V}}
\newcommand{\hE}{\hat{E}}
\newcommand{\hn}{\hat{n}}
\newcommand{\hgG}{\hat{\gG}}

\theoremstyle{definition}
\newtheorem{definition}{Definition}[section]

\newcommand{\ptitlenoskip}[1]{\noindent{\bf #1.}}

\newcommand{\acronym}{FONDUE}
\usepackage{times}

\title{FONDUE: A Framework for Node Disambiguation Using Network Embeddings}

\author{ {\bf Ahmad Mel} \hspace{10pt} {\bf Bo Kang \hspace{10pt} {\bf Jefrey Lijffijt} \hspace{10pt} {\bf Tijl De Bie} } \\
IDLab, Ghent University\\
\emph{firstname.lastname@ugent.be}
}

\begin{document}

\maketitle

\begin{abstract}
  Real-world data often presents itself in the form of a network.
  Examples include social networks, citation networks, biological networks, and knowledge graphs.
  In their simplest form, networks represent real-life entities (e.g. people, papers, proteins, concepts) as nodes,
  and describe them in terms of their relations with other entities by means of edges between these nodes.
  This can be valuable for a range of purposes from the study of information diffusion
  to bibliographic analysis, bioinformatics research, and question-answering.

  The quality of networks is often problematic though, affecting downstream tasks.
  This paper focuses on the common problem where a node in the network
  in fact corresponds to multiple real-life entities.
  In particular, we introduce \acronym{},
  an algorithm based on network embedding for node disambiguation.
  Given a network, \acronym{} identifies nodes that correspond to multiple entities,
  for subsequent splitting.
  Extensive experiments on twelve benchmark datasets demonstrate that \acronym{} is substantially and uniformly more accurate for ambiguous node identification compared to the existing state-of-the-art, at a comparable computational cost, while less optimal for determining the best way to split ambiguous nodes.

\end{abstract}

\section{Introduction}\label{sec:intro}

Increasingly, data naturally takes the form of a network of interrelated entities.
Examples include social networks describing social relations between people (e.g. Facebook),
citation networks describing the citation relations between papers (e.g. DBLP),
biological networks e.g. describing interactions between proteins (e.g. DIP),
and knowledge graphs describing relations between concepts or objects (e.g. DBPedia).
Thus new machine learning, data mining, and information retrieval methods
are increasingly targeting data in their native network representation.

An important problem across all of data science, broadly speaking, is data quality.
For problems on networks, especially those that are successful in exploiting
fine- as well as coarse-grained structure of networks,
ensuring good data quality is perhaps even more important than in standard tabular data.
For example, an incorrect edge can have a dramatic effect on the implicit representation of other nodes,
by dramatically changing distances on the network.
Similarly, mistakenly representing distinct real-life entities by the same node in the network
may dramatically alter its structural properties.

\paragraph{The Node Disambiguation problem}
While identifying missing edges, and conversely, identifying incorrect edges,
can be tackled adequately using link prediction methods,
prior work has neglected the other task: identifying nodes that are ambiguous---%
i.e. nodes that correspond to more than one real-life entity.
We will refer to this task as Node Disambiguation (ND).

\begin{figure*}[h]
  \centering
      \includegraphics[width=\textwidth]{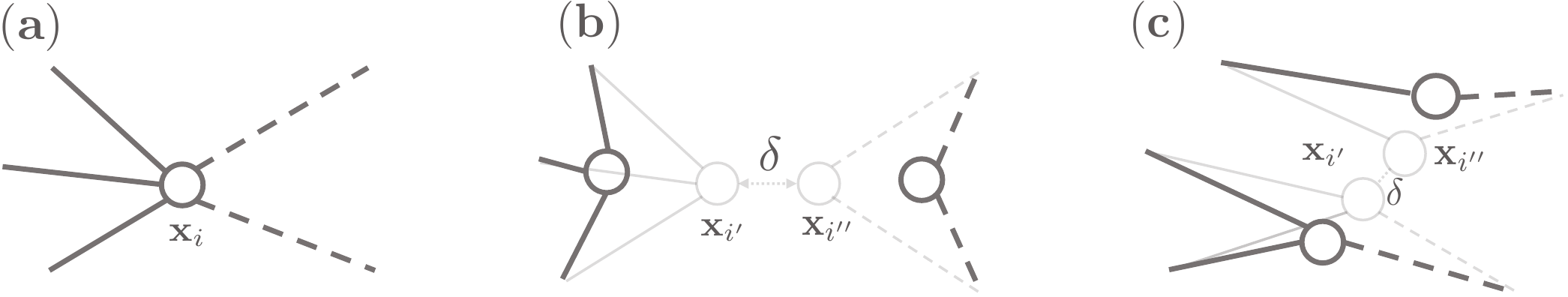}
      \caption{(a) A node that corresponds to two real-life entities that belongs to two communities. Links that connect the node with different communities are plotted in either full lines or dashed lines. (b) an ideal split that aligns well with the communities. (c) a less optimal split.}
      \label{fig:illustration}
\end{figure*}

In this paper, we address the problem of ND in the most basic setting:
given a network, unweighted, unlabeled, and undirected,
the task considered is to identify nodes that correspond to multiple distinct real-life entities.
We formulate this as an inverse problem,
where we want to use the \emph{ambiguous network} (which contains ambiguous nodes)
in order to retrieve the \emph{unambiguous network} (in which all nodes are unambiguous).
Clearly, this inverse problem is ill-posed, making it impossible to solve
without additional information, a prior, or another type of inductive bias.

Such an inductive bias can be provided by the Network Embedding (NE) literature,
which has produced embedding-based models
that are capable of accurately modeling the connectivity of \emph{real-life} networks down to the node-level,
while being unable to accurately model \emph{random} networks~\citep{wang2017community, fortunato2007resolution}.
Inspired by this literature,
we propose to use as an inductive bias the fact that the unambiguous network must be easy to model using a NE.
Thus, we introduce FONDUE (a Framework fOr Node Disambiguation Using network
Embeddings), a method that determines the extent to which
splitting a node into distinct entities would improve the quality of the resulting NE.

\paragraph{Example}~\label{sec:example} Figure~\ref{fig:illustration}(a) illustrates the idea of FONDUE applied on a single node $\vx_i$. In this example, node $\vx_i$ corresponds to two real-life entities that belong to two separate communities, visualized by either full or dashed lines, to highlight the distinction.
% The links that connect this ambiguous node with nodes from two communities are visualized in either full line or dashed line.
Because node $\vx_i$ is connected to two different communities,
this should be reflected in the embedding space, given that network embeddings do model this relationship, thus $\vx_i$ would be naturally located
% it should be situated close to nodes in both communities in the embedding, therefore to achieve both goals, in the node sitting in
between both communities. Figure~\ref{fig:illustration}(b) shows an ideal split where the two resulting nodes $\vx_{i'}$ and $\vx_{i''}$ are embedded close to their own respective community. Figure~\ref{fig:illustration}(c) shows a less ideal split where the two resulting nodes are still embedded in the middle of their two distinct communities.

\paragraph{Related work}~\label{sec:relwork}
The problem of ND differs from Named-Entity Disambiguation
(NED; also known as Named Entity Linking, NEL),
a Natural Language Processing (NLP) task where the purpose is to identify which real-life entity from a list
a named-entity in a text refers to.
For example, in the ArnetMiner Dataset~\citep{Tang:12TKDE} `Bin Zhu' corresponds to more than 10+ authors.
NED in this context aims to match the author names to unique (unambiguous)
author identifiers~\citep{Tang:12TKDE, shen_entity_2015, zhang_name_2017, parravicini_fast_2019}.
NED typically strongly relies on the text, e.g. by characterizing the context in which the named entity occurs (e.g. paper topic).
In ND, in contrast, no natural language is considered,
and the goal is to rely on just the network's connectivity
in order to identify which nodes may correspond to multiple distinct entities.
Moreover, ND does not assume the availability of a list of known unambiguous entity identifiers,
such that an important part of the challenge is to identify which nodes are ambiguous in the first place.%
\footnote{Note that ND could be used to assist in NED tasks,
e.g. if the natural language is used to create a graph of related named entities.
This is left for further work.}

The research by~\citep{saha2015name,hermansson2013entity} is most closely related to ours.
These papers also only use topological information of the network for ND.
Yet, \citep{saha2015name} also require timestamps for the edges,
while \citep{hermansson2013entity} require a training set of nodes labeled as ambiguous and non-ambiguous.
Moreover, even though the method proposed by \citep{saha2015name} is reportedly orders of magnitude faster than the one proposed by \citep{hermansson2013entity},
it remains computationally substantially more demanding than \acronym{} (e.g. \citep{saha2015name} evaluate their method on networks with just 150 entities).
Other recent work using NE for NED~\citep{zhang2017name,xu2018network, chen2017task, 10.1145/3132847.3132925}
is only related indirectly as they rely on additional information besides the topology of the network.

\paragraph{Contributions}
In this paper we propose FONDUE, which exploits the fact that naturally occurring networks can be embedded well
using state-of-the-art NE methods,
by identifying nodes as more likely to be ambiguous if splitting them enhances the quality of an optimal NE.
To do this in a scalable manner, substantial challenges had to be overcome.
Specifically, through a first-order analysis we derive a fast approximation
of the expected NE quality improvement after splitting a node.
We implemented this idea for CNE~\citep{kang2019conditional}, a recent state-of-the-art NE method,
although we demonstrate that the approach can be applied for a broad class of NE methods.
Our extensive experiments over a wide range of networks demonstrate the superiority of \acronym{} in comparison with the best available baselines for ambiguous node \emph{identification},
and this at comparable computational cost.
We also empirically demonstrate that, somewhat surprisingly, this increase in identification accuracy is not matched by a comparable improvement in ambiguous node \emph{splitting} accuracy.
Thus, we recommend using \acronym{} for ambiguous node identification in combination with a state-of-the-art approach for optimally splitting the identified nodes.

\section{Methods}\label{sec:methods}

Section~\ref{sec:problem} formally defines the ND problem.
Section~\ref{sec:approach} introduces the \acronym{} approach in a generic manner, independent of the specific NE method it is applied to.
A scalable approximation of \acronym{} is described in Section~\ref{sec:scale}.
Section~\ref{sec:cne} then develops \acronym{} in detail for the embedding method CNE~\citep{kang2019conditional}.

%Given an original graph that is nicely embeddable, that is somehow altered by data gathering process (parsing, etc\dots) into a contracted graph (namesakes),

\ptitlenoskip{Notation}
Throughout this paper, a bold uppercase letter denotes matrix (e.g. $\mA$),
bold lower case letter denotes a column vector (e.g. $\vx_i$), $\tran{(.)}$
denotes matrix transpose (e.g. $\tran{\mA}$), and $\frob{(.)}$ denotes the
Frobenius norm of a matrix (e.g. $\frob{\mA}$).
%$(.)^*$ denotes the optimal value of a variable w.r.t. a specified optimization problem.

\subsection{Problem definition}\label{sec:problem}

We denote an undirected, unweighted, unlabeled graph as $\gG = (V, E)$,
with $V=\{ 1,2,\ldots,n\}$ the set of $n$ nodes (or vertices),
and $E \subseteq \binom{V}{2}$ the set of edges (or links)
between these nodes.
We also define the adjacency matrix of a graph $\gG$,
denoted $\mA \in \{0,1\}^{n\times n}$, as $\emA_{ij}=1$ iff $\lij \in E$.
We denote $\va_i \in \{0, 1\}^n$ as the adjacency vector for node $i$,
i.e. the $i$th column of the adjacency matrix $\mA$,
and $\Gamma(i) = \{ j \pp \lij \in E \}$ the set of neighbors of $i$.

% \begin{definition}[Ambiguous node]\label{def:eq}
%
% \emph{Given $\mA$ the adjacency matrix of $\gG= (V, E)$, and $\va_i$ the $i$th row vector
% of $\mA$, and
% we define an ambiguous node $v_i \in {V}$, and resolved nodes $v_{\amb{i}},  v_{\ambb{i}}$
% with their respective row vectors $\va_{\amb{i}}, \va_{\ambb{i}} \in \{0, 1\}^{1\times n}$ every node $i$ satisfying the following properties:}
%
% \begin{itemize}
%   \item $\Gamma(v_{\amb{i}}) \cup \Gamma(v_{\ambb{i}}) = \Gamma(v_{i})$
%   \item $\Gamma(v_{\amb{i}}) \cap \Gamma(v_{\ambb{i}}) = \emptyset$
%   \item ${\va_i} =  \va_{\amb{i}} +  \va_{\ambb{i}}$
% \end{itemize}
% \end{definition}

To formally define the ND problem as an inverse problem,
we first need to define the forward problem which maps an unambiguous graph onto an ambiguous one.
To this end, we define a node contraction:
\begin{definition}[Node Contraction]\label{def:contraction}
A node contraction $c$ for a graph $\gG=(V,E)$ with $V=\{1,2,\ldots,n\}$
is a surjective function $c:V\rightarrow \hV$
for some set $\hV=\{1,2,\ldots,\hn\}$ with $\hn\leq n$.
For convenience, we will define $c^{-1}:\hV\rightarrow 2^V$ as $c^{-1}(i)=\{k\in V|c(k)=i\}$ for any $i\in\hV$.
Moreover, we will refer to the cardinality $|c^{-1}(i)|$ as the \emph{multiplicity} of the node $i\in\hV$.
\end{definition}
A node contraction defines an equivalence relation $\sim_c$ over the set of nodes:
$i\sim_c j$ iff $c(i)=c(j)$,
and the set $\hV$ is the quotient set $V/\sim_c$.
The contraction can thus be used to define the concept of an ambiguous graph, as follows.
\begin{definition}[Ambiguous graph]\label{def:cont_graph}
Given a graph $\gG=(V,E)$ and a node contraction $c$ for that graph,
the ambiguous graph $\hgG$ is defined as $\hgG=(\hV,\hE)$
where $\hE=\{\{i,j\}|\exists \{k,l\}\in E:c(k)=i\land c(l)=j\}$.
Overloading notation, we write $\hgG\triangleq c(\gG)$.
We refer to $\gG$ as the unambiguous graph.
\end{definition}

We can now formally define the ND problem as inverting the contraction operation:
\begin{definition}[The Node Disambiguation Problem]\label{def:nd}
Given an ambiguous graph $\hgG=(\hV,\hE)$ (denoted using hats to indicate this is typically the empirically observed graph),
ND aims to retrieve the unambiguous graph $\gG=(V,E)$
and associated node contraction $c$,
i.e. for which $c(\gG)=\hgG$.
\end{definition}
To be clear, it suffices to identify $\gG$ up to an isomorphism,
as the actual identifiers of the nodes are irrelevant.
Equivalently, it suffices to identify the multiplicities of all nodes $i\in\hgG$,
i.e. the number of unambiguous nodes that each node in the ambiguous graph represents.
The actual node identifiers in $c^{-1}(i)$ are irrelevant.

Clearly, the ND problem is an ill-posed inverse problem.
Thus, further assumptions, inductive bias, or priors are inevitable in order to solve the problem.

The key idea in \acronym{} is that $\gG$, considering it is a `natural' graph,
can be embedded well using state-of-the-art NE methods (which have empirically been shown to embed `natural' graphs well).
Thus, \acronym{} searches for the graph $\gG$ such that $c(\gG)=\hgG$,
while optimizing the NE cost function.

Without loss of generality, the ND problem can be decomposed into two steps:
\begin{enumerate}
\item Estimating the multiplicities of all $i\in\hgG$---%
i.e. the number unambiguous nodes from $\gG$ represented by the node from $\hgG$.
Note that the number of nodes $n$ in $V$ is then equal to the sum of these multiplicities,
and arbitrarily assigning these $n$ nodes to the sets $c^{-1}(i)$ defines $c^{-1}$ and thus $c$.
\item Given $c$, estimating the edge set $E$.
To ensure that $c(\gG)=\hgG$, for each $\lij\in\hE$ there must exist at least one edge
$\lkl\in E$ with $k\in c^{-1}(i)$ and $l\in c^{-1}(j)$.
\end{enumerate}
As an inductive bias for the second step, we will additionally assume that the graph $\gG$ is sparse.
Thus, \acronym{} estimates $\gG$ as the graph with the smallest set $E$ for which $c(\gG)=\hgG$.
Practically, this means that an edge $\lij\in\hE$ results in exactly one edge $\lkl\in E$ with $k\in c^{-1}(i)$ and $l\in c^{-1}(j)$,
and that equivalent nodes $k\sim_c l$ with $k,l\in V$ are never connected by an edge, i.e. $\lkl\not\in E$.
This bias is justified by the sparsity of most `natural' graphs, and our experiments indicate it is justified.

\subsection{\acronym{} as a generic approach}\label{sec:approach}
To address the ND problem~\ref{def:nd}, \acronym{} uses an inductive bias that the unambiguous network must be easy to model using NE. This allows us to approach the ND problem in the context of NE. Here we first introduce the basic concepts in NE and then present \acronym{}.

\ptitlenoskip{Network Embedding} NE methods find a mapping $f: V \to \sR^d$ from nodes to $d$-dimensional real vectors. An embedding is denoted as $\mX = \tran{(\vx_1, \vx_2, \ldots, \vx_n)} \in \sR^{n\times d}$, where $\vx_i \triangleq f(i)$ for $i \in V$ is the embedding of each node.
All well-known NE methods aim to find an optimal embedding $\mX^*_{\gG}$ for given graph $\gG$ that minimizes a continuous differentiable cost function $\cO(\gG, \mX)$.

Thus, based on NE, the ND problem~\ref{def:nd} can be restated as follows.
\begin{definition}[NE based ND problem]\label{def:ne_nd}
Given an ambiguous graph $\hgG$, NE based ND aims to retrieve the unambiguous graph $\gG$ and the associated contraction $c$:
\begin{align*}
  \underset{\gG}{\text{argmin}}~\ \ &\cO\left(\gG, \mX^*_{\gG}\right) \\
  \text{s.t.}\ \ & c(\gG) = \hgG.
\end{align*}
\end{definition}
Ideally, this optimization problem can be solved by simultaneously finding optimal splits for all nodes (i.e., a reverse mapping) that yield the smallest embedding cost after re-embedding. However, this strategy requires to (a) search splits in an exponential search space that has the combinations of splits (with arbitrary cardinality) of all nodes, (b) to evaluate each combination of the splits, the embedding of the resulting network needs to be recomputed. Thus, this ideal solution is computationally intractable and more scalable solutions are needed.

\ptitlenoskip{\acronym} We approach the NE based ND problem~\ref{def:ne_nd} in a greedy and iterative manner. At each iteration, \acronym{} identifies the node that has a split which will result in the smallest value of the cost function among all nodes. To further reduce the computational complexity, \acronym{} only split one node into two nodes at a time (e.g. Figure~\ref{fig:illustration}(b)), i.e., it splits node $i$ into two nodes $i'$ and $i''$ with corresponding adjacency vectors $\va_{i'}, \va_{i''} \in \{0,1\}^n$, $\va_{i'} + \va_{i''} = \va_{i}$. For convenience, we refer to such a split as a \emph{binary split}. Once, the best binary split of the best node is identified, \acronym{} splits that node and starts the next iteration.

% This substantially reduces the complexity from enumerating the combinations of arbitrary splits among all nodes to enumerating binary splits of one node for each node. As discussed in the next subsection, this iterative approach allows approximation search strategies to be applied.

However, the evaluation of each split requires recomputing the embedding, which is still computationally demanding. Instead of recomputing the embedding, \acronym{} performs a first-order analysis by investigates the effect of an infinitesimal split of a node $i$ around its embedding $\vx_i$, on the cost $O(\hgG_{si}, \hmX_{si})$ obtained after performing the splitting. Specifically, \acronym{} seeks the split of node $i$ that will result in embedding $\vx_{i'}$ and $\vx_{i''}$ with infinitesimal difference $\vdelta_i$ (where $\vdelta_i = \vx_{i'} - \vx_{i''}$, $\vx_{i'}=\vx_i+\frac{\vdelta_i}{2}$, $\vx_{i''}=\vx_i-\frac{\vdelta_i}{2}$, and $\vdelta_i \to \vZero$, e.g. Figure~\ref{fig:illustration}(b)) such that  $||\nabla_{\vdelta_i}\cO(\hgG_{si}, \hmX_{si})||$ is large. This can be done analytically. Indeed, applying the chain rule, we find:
\begin{align}\label{eq:inf_grad_sum}
  &\nabla_{\vdelta_i}\cO(\hgG_{si}, \hmX_{si}) \nonumber \\
  &= \nabla_{\vx_{i'}}\cO(\hgG_{si}, \hmX_{si})\cdot\nabla_{\vdelta_i}\vx_{i'} \nonumber + \nabla_{\vx_{i''}}\cO(\hgG_{si}, \hmX_{si})\cdot\nabla_{\vdelta_i}\vx_{i''} \nonumber \\
  &= \frac{1}{2} \nabla_{\vx_{i'}}\cO(\hgG_{si}, \hmX_{si}) - \frac{1}{2} \nabla_{\vx_{i''}} \cO(\hgG_{si}, \hmX_{si})
\end{align}
We can continue the derivation by further realizing that random-walk based and probabilistic model based NE methods like node2vec~\citep{grover2016node2vec}, LINE~\citep{tang2015line}, CNE~\citep{kang2019conditional}, aim to embed similar nodes in the graph closer to each other and vice versa. Thus their objective functions can be decomposed as:
\begin{align*}
  \cO(\gG, \mX) &= \sum_{j:\lij \in E}\cO^p(\mA_{ij}=1, \vx_i, \vx_j) \\
  &+ \sum_{l:\lkl \notin E}\cO^p(\mA_{kl}=0, \vx_k, \vx_l),
\end{align*}
where $\cO^p(\mA_{ij}=1, \vx_i, \vx_j)$ is the part of objective function that corresponds to node $i$ and node $j$ with an edge between them ($\mA_{ij}=1$) and $\cO^p(\mA_{kl}=0, \vx_k, \vx_l)$ is the part of objective function, where node $k$ and node $l$ are disconnected.

Denote $\mF^1_i \in \sR^{d\times |\Gamma(i)|}$ as the matrix with the $j$-th column corresponds to gradient $\nabla_{\vx_i}\cO^p(\mA_{ij}=1, \vx_i, \vx_j)$ and $j \in \Gamma(i)$, $\mF^0_i \in \sR^{d\times |\Gamma(i)|}$ as the matrix with the $l$-th column corresponds to gradient $\nabla_{\vx_i}\cO^p(\mA_{il}=0, \vx_i, \vx_l)$ and $l \in \Gamma(i)$. Let $\vb_i \in \{1,-1\}^{|\Gamma_i|}$ to be a vector where each element corresponds to a neighbor of $i$, the ``$1$'' elements correspond to the neighobrs of $i'$ and ``$-1$'' elements correspond to the neighbors of $i''$. Then the gradient Eq.~\ref{eq:inf_grad_sum} can be further derived as
\begin{align}\label{eq:m_gradient}
  \nabla_{\vdelta_i}\cO(\hgG_{si}, \hmX_{si}) &=  \frac{1}{2}(\mF_i^1 - \mF_i^0)\vb_i
\end{align}

Denote $\vb_i = \va_{i'} - \va_{i''}$, and
\begin{equation}\label{eq:m_matrix}
  \mM_i = \tran{(\mF_i^1 - \mF_i^0)}(\mF_i^1 - \mF_i^0),
\end{equation}
the goal that \acronym{} aims to achieve can be summarized in are more compact form:
\begin{align}\label{eq:quotient}
  \underset{i, \vb_i}{\text{argmax}} \frac{\tran{\vb_i}\mM_i\vb_i}{\tran{\vb_i}\vb_i}
\end{align}
Note that $\mM_i\succeq \vZero$ for all nodes and all splits, such that this is an instance of Boolean Quadratic Maximization problem \citep{nesterov1998semidefinite,luo2010semidefinite}. This problem is NP-hard, thus we need an approximation solution.

\subsection{Making \acronym{} scale}\label{sec:scale}
In order to efficiently search for best split on a given node, we developed two approximation heuristics.

First, we randomly split the neighborhood $\Gamma(i)$ into two and evaluate the objective Eq.~\ref{eq:quotient}. Repeat the randomization procedure for a fixed number of times, pick the split that gives the best objective value as output.

Second, we find the eigenvector $\vv$ that corresponds to the largest absolute eigenvalue of matrix $\mM_i$. Sort the element in vector $\vv$ and assigning top $k$ corresponding nodes to $\Gamma(i')$ and the rest to $\Gamma(i'')$. evaluating the objective value for $k=1 \ldots |\Gamma(i)|$ and pick the best split.

Finally, we combine theses two heuristics and use the split that gives best objective Eq.~\ref{eq:quotient} as the final split of the node $i$.

\subsection{\acronym{} using CNE}\label{sec:cne}
We now apply \acronym{} to Conditional Network Embedding (CNE). CNE proposes a probability distribution for network embedding and finds a locally optimal embedding by maximum likelihood estimation. CNE has objective function:
\begin{align*}
&\cO(\gG,\mX)=\ \log(P(\mA|\mX))\\
&=\sum_{ij:\mA_{ij}=1} \log{P_{ij}(\mA_{ij}=1|\mX)} \\
&+ \sum_{ij:\mA_{ij}=0} \log{P_{ij}(\mA_{ij}=0|\mX)}.
\end{align*}
Here, the link probabilities $P_{ij}$ conditioned on the embedding are defined as follows:
\begin{align*}
&P_{ij}(\mA_{ij}=1|\mX) = \\
&\frac{P_{\mA,ij}\cN_{+,\sigma_1}(\|\vx_i-\vx_j\|)}{P_{\mA,ij}\cN_{+,\sigma_1}(\|\vx_i-\vx_j\|) + (1-P_{\mA,ij})\cN_{+,\sigma_2}(\|\vx_i-\vx_j\|)}, \nonumber
\end{align*}
where $\cN_{+,\sigma}$ denotes a half-Normal distribution \citep{leone1961folded} with spread parameter $\sigma$, $\sigma_2>\sigma_1=1$, and where $P_{\hmA,ij}$ is a prior probability for a link to exist between nodes $i$ and $j$ as inferred from the degrees of the nodes (or based on other information about the structure of the network \citep{van2016subjective}).
In order to compute the \acronym{} objective Eq.~\ref{eq:quotient}, first we derive the gradient:
\begin{align*}
  &\nabla_{\vx_i}\cO(\gG, \mX) \\
  &= \gamma\sum_{j:\lij \in E} (\vx_i-\vx_j)\left(P\left(\mA_{ij}=1|\mX\right)-1\right) \\
  &+ \gamma\sum_{l:\lil \notin E} (\vx_i-\vx_l)\left(P\left(\mA_{il}=1|\mX\right)-0\right).
\end{align*}
where $\gamma = \frac{1}{\sigma_1^2} - \frac{1}{\sigma_2^2}$. Then the $j$-th column of term $\mF^1_i - \mF^0_i$ in gradient Eq.~\ref{eq:m_gradient} reads
\begin{align*}
  \left(\mF^1_i - \mF^0_i\right)_{:,j} &= \gamma (\vx_i-\vx_j)\left(P\left(\mA_{ij}=1|\mX\right)-1\right) \nonumber \\
  &- \gamma (\vx_i-\vx_j)\left(P\left(\mA_{ij}=1|\mX\right)\right) \nonumber\\
  &= -\gamma(\vx_i-\vx_j)
\end{align*}
This allows us to further compute vectorized gradient Eq.~\ref{eq:m_gradient}:
\begin{align*}
  \nabla_{\vdelta_i}\cO(\hgG_{si}, \hmX_{si}) &=  -\frac{\gamma}{2}
  \begin{pmatrix}
    \begin{array}{ccc}
    \vdots & \vx_i - \vx_j & \vdots
    \end{array}
\end{pmatrix}
\vb_i
\end{align*}
Now we can compute matrix $\mM_i$:
\begin{align*}
  \mM_i &= \tran{(\mF_i^1 - \mF_i^0)}(\mF_i^1 - \mF_i^0) \\
  &= \gamma^2\sum_{k,l\in\Gamma_i}(\vx_i - \vx_k)\tran{(\vx_i - \vx_l)}
\end{align*}
Plug $\mM_i$ into Eq.~\ref{eq:quotient}, and omitting the constant factor ${\gamma^2}$, the Boolean Quadratic Maximization problem based on CNE has the following form:
\begin{align}\label{eq:obj_cne}
  \underset{i, \vb_i}{\text{argmax}} \frac{\tran{\vb_i}\sum_{k,l\in\Gamma{i}}(\vx_i - \vx_k)\tran{(\vx_i - \vx_l)}\vb_i}{\tran{\vb_i}\vb_i}
\end{align}

\begin{table}[]
    \begin{adjustbox}{width=\columnwidth,center}
      \begin{tabular}{@{}cl@{}}
      \toprule
      \textbf{Datasets} & \multicolumn{1}{c}{\textbf{Description}} \\ \midrule
      \textbf{FB-SC} & \begin{tabular}[c]{@{}l@{}}Facebook Social Circles network~\citep{snapnets} \\ consists of anonymized friends list from Facebook\end{tabular} \\ \midrule
      \textbf{FB-PP} & \begin{tabular}[c]{@{}l@{}}Page-Page graph of verified Facebook pages~\citep{snapnets}. \\ Nodes represent official Facebook pages while the \\ links are mutual likes between pages.\end{tabular} \\ \midrule
      \textbf{email} & \begin{tabular}[c]{@{}l@{}}Anonymized network generated using email data from a large \\ European research institution modelling the incoming and \\ outgoing email exchange between its members.~\citep{snapnets}\end{tabular} \\ \midrule
      \textbf{STD} & \begin{tabular}[c]{@{}l@{}}A network of student database of the Computer Science department of the\\ University of Antwerp that represent the connections between students, \\ professors and courses.~\citep{goethals2010mining}\end{tabular} \\ \midrule
      \textbf{PPI} & \begin{tabular}[c]{@{}l@{}}A subnetwork of the BioGRID Interaction Database\\ ~\citep{breitkreutz2007biogrid}, that uses PPI network for Homo Sapiens.\end{tabular} \\ \midrule
      \textbf{lesmis} & \begin{tabular}[c]{@{}l@{}}A network depicting the coappearance of characters in \\ the novel Les Miserables.\citep{knuth1993stanford}\end{tabular} \\ \midrule
      \textbf{netscience} & \begin{tabular}[c]{@{}l@{}}A coauthorship network of scientists working on network theory and \\ experiment.~\citep{snapnets}\end{tabular} \\ \midrule
      \textbf{polbooks} & \begin{tabular}[c]{@{}l@{}}Network of books about US politics, with edges between books \\ represent frequent copurchasing of books by the same buyers.\footnote{http://www-personal.umich.edu/~mejn/netdata/}\end{tabular} \\ \midrule
      \textbf{GrQc} & Collaboration network of Arxiv General Relativity~\citep{newman2001structure} \\ \midrule
      \textbf{CondMat03} & Collaboration network of Arxiv Condensed Matter till 2003~\citep{newman2001structure} \\ \midrule
      \textbf{CondMat05} & Collaboration network of Arxiv Condensed Matter till 2005~\citep{newman2001structure} \\ \midrule
      \textbf{AstroPh} & Collaboration network of Arxiv Astro Physics~\citep{newman2001structure} \\ \bottomrule
      \end{tabular}
\end{adjustbox}
\caption{The different datasets used in our experiments. (Sec.~\ref{sec:dataset}). }
\label{tab:dataset}
\end{table}

\begin{table*}[ht]
  \caption{Various properties about each network used in our experiments}
  \begin{adjustbox}{width=\textwidth, center}
    \begin{tabular}{lcccccccccccc}
    \hline
     & \textbf{Email} & \textbf{PPI} & \textbf{GrQc} & \textbf{lesmis} & \textbf{netscience} & \textbf{polbooks} & \textbf{FB-SC} & \textbf{FB-PP} & \textbf{STD} & \textbf{AstroPh} & \textbf{CM05} & \textbf{CM03} \\ \hline
    \textbf{\# Nodes} & 986 & 3,852 & 4,158 & 77 & 379 & 105 & 4,039 & 22,470 & 395 & 14,845 & 36,458 & 27,519 \\
    \textbf{\# Edges} & 16,687 & 38,705 & 13,428 & 254 & 914 & 441 & 88,234 & 171,002 & 3,423 & 119,652 & 171,735 & 116,181 \\
    \textbf{Avg degree} & 33.8 & 20.1 & 6.5 & 6.6 & 4.8 & 8.4 & 43.7 & 15.2 & 17.3 & 16.1 & 9.4 & 8.4 \\
    \textbf{Density} & 3E-02 & 5E-03 & 2E-03 & 9E-02 & 1E-02 & 8E-02 & 1E-02 & 7E-04 & 4E-02 & 1E-03 & 3E-04 & 3E-04 \\ \hline
    \end{tabular}
\end{adjustbox}
\label{tab:stat}
\end{table*}

\begin{table*}[ht]
\caption{Performance evaluation (AUC score) on multiple datasets for our methods compared with other baselines. Note that for some of the datasets with small number of nodes, we did not perform any contraction for $0.001$ as the number of contracted nodes in this case is very small, thus we replaced the values for those methods by "-".}
\begin{adjustbox}{width=\textwidth,center}
  \begin{tabular}{@{}|l|cccc|cccc|cccc|@{}}
  \toprule
  \multicolumn{1}{|c|}{\textbf{\% of ambiguous nodes}} & \multicolumn{4}{c|}{\textbf{0.1\%}} & \multicolumn{4}{c|}{\textbf{1\%}} & \multicolumn{4}{c|}{\textbf{10\%}} \\ \midrule
  \multicolumn{1}{|c|}{\textbf{Methods}} & \multicolumn{1}{l}{\textbf{FONDUE}} & \multicolumn{1}{l}{\textbf{NC}} & \multicolumn{1}{l}{\textbf{CC}} & \multicolumn{1}{l|}{\textbf{degree}} & \multicolumn{1}{l}{\textbf{FONDUE}} & \multicolumn{1}{l}{\textbf{NC}} & \multicolumn{1}{l}{\textbf{CC}} & \multicolumn{1}{l|}{\textbf{degree}} & \multicolumn{1}{l}{\textbf{FONDUE}} & \multicolumn{1}{l}{\textbf{NC}} & \multicolumn{1}{l}{\textbf{CC}} & \multicolumn{1}{l|}{\textbf{degree}} \\ \midrule
  \textbf{email} & \textbf{-} & - & - & - & \textbf{0.747} & 0.600 & 0.402 & 0.702 & \textbf{0.728} & 0.507 & 0.311 & 0.712 \\
  \textbf{ppi} & \textbf{0.775} & 0.516 & 0.623 & 0.729 & \textbf{0.737} & 0.495 & 0.643 & 0.723 & \textbf{0.727} & 0.522 & 0.628 & 0.716 \\
  \textbf{lesmis} & - & - & - & - & - & - & - & - & \textbf{0.799} & 0.513 & 0.412 & 0.733 \\
  \textbf{netscience} & - & - & - & - & \textbf{0.918} & 0.839 & 0.802 & 0.818 & \textbf{0.897} & 0.841 & 0.720 & 0.792 \\
  \textbf{polbooks} & - & - & - & - & \textbf{0.836} & 0.680 & 0.598 & 0.755 & \textbf{0.868} & 0.716 & 0.329 & 0.820 \\
  \textbf{FB-SC} & \textbf{0.933} & 0.849 & 0.548 & 0.743 & \textbf{0.939} & 0.779 & 0.446 & 0.745 & \textbf{0.915} & 0.807 & 0.158 & 0.749 \\
  \textbf{FB-PP} & \textbf{0.905} & 0.722 & 0.738 & 0.715 & \textbf{0.890} & 0.727 & 0.745 & 0.723 & \textbf{0.871} & 0.722 & 0.742 & 0.722 \\
  \textbf{STD} & - & - & - & - & \textbf{0.740} & 0.438 & 0.466 & 0.701 & \textbf{0.712} & 0.574 & 0.528 & 0.710 \\
  \textbf{GrQc} & \textbf{0.852} & 0.815 & 0.813 & 0.759 & \textbf{0.854} & 0.805 & 0.799 & 0.739 & \textbf{0.846} & 0.809 & 0.789 & 0.743 \\
  \textbf{condmat05} & \textbf{0.880} & 0.845 & 0.823 & 0.746 & \textbf{0.881} & 0.852 & 0.815 & 0.750 & \textbf{0.865} & 0.852 & 0.813 & 0.755 \\
  \textbf{condmat03} & \textbf{0.891} & 0.850 & 0.823 & 0.745 & \textbf{0.881} & 0.849 & 0.820 & 0.759 & \textbf{0.864} & 0.849 & 0.812 & 0.758 \\
  \textbf{AstroPh} & \textbf{0.865} & 0.824 & 0.769 & 0.724 & \textbf{0.857} & 0.833 & 0.780 & 0.730 & \textbf{0.837} & 0.836 & 0.758 & 0.731 \\ \bottomrule
  \end{tabular}
\end{adjustbox}
\label{tab:quant}
\end{table*}

\section{Experiments}\label{sec:experiments}
% In this section we evaluate the performance of our method on multiple datasets~\ref{sec:dataset}, for solving the 2 steps of the ND problem, ie. identifying the set of ambiguous nodes $\hV$ given an ambiguous (contracted) graph $\hgG$, then second task would be finding the optimal edge reassignment for each ambiguous nodes $\hat{v} \in \hV$ (node split).

In this section, we investigate the following questions: $\textbf{Q}_1$ Quantitatively, how does our method perform in identifying ambiguous nodes compared to the state-of-the-art and other heuristics? (Sec.~\ref{sec:identifying});
 $\textbf{Q}_2$ Quantitatively, how does our method perform in terms of splitting the ambiguous nodes? (Sec.~\ref{sec:splitting}); $\textbf{Q}_3$ How does the behavior of the method change when the degree of contraction of a network varies? (Sec.~\ref{sec:param-sens}); $\textbf{Q}_4$ Does the proposed method scale? (Sec.~\ref{sec:runtime}).

 \subsection{Datasets}\label{sec:dataset}
 One main challenge for assessing the evaluation of disambiguation tasks is the the scarcity of availability of ambiguous (contracted) graph datasets with reliable ground truth.
 Thus we opted to create a contracted graph given a source graph, and then use the latter as ground truth to assess the accuracy of our method.

 % This is done by applying random node contraction on each of the networks in the datasets.

 % \subsection{Preprocessing: Network Contraction}\label{sec:node-contrac}
 More specifically, for each network $\gG=(V,E)$, a graph contraction was performed to create a contracted graph $\hgG=(\hat{V},\hat{E})$ (ambiguous) by randomly merging a fraction $r$ of total number of nodes, to create a ground truth to test our proposed method. This is done by first specifying the fraction of the nodes in the graph to be contracted ($r \in \{0.001, 0.01, 0.1\}$), and then sampling two sets of vertices, $\hat{V}^i \subset \hat{V}$ and $\hat{V}^j \subset \hat{V}$, such that
 $|\hat{V}^i| = |\hat{V}^j| = \floor{r\cdot|\hat{V}|}$ and $\hat{V}^i \cap \hat{V}^j = \emptyset$.
 Then, every element
 $v_j\in \hat{V}^j$
 is merged into the corresponding $v_i \in \hat{V}^i$ by reassigning the links connected to $v_j$ to $v_i$ and removing $v_j$ from the network. The node pairs $(v_i, v_j)$ later serve as ground truths.

 We've tested the performance of \acronym{}, as well as that of the competing methods, on 12 different datasets listed in Table~\ref{tab:dataset}, with their properties shown in Table~\ref{tab:stat}.
 % The DBLP dataset was used for qualitative evaluation of our method, as it's the only graph that contains ground truth, i.e. different authors with the same name. Note that
 % the dataset itself was quite noisy, and preprocessing was necessary to validate the ground truth labels. This was done by scraping the DBLP website for different author names that have published papers in machine learning venues (NeurIPS, ICML, ICLR, ECML/PKDD), resulting in $47$ ambiguous names for a network of size $15,277$.

\subsection{Quantitative Evaluation of Node Identification}\label{sec:identifying}
In this section, we focus on answering $\textbf{Q}_1$, namely, given a contracted graph, \acronym{} aims to identify the list of contracted (ambiguous) nodes present in it.

\ptitlenoskip{Baselines}\label{sec:baselinez}
As mentioned earlier in Sec.~\ref{sec:intro}, most entity disambiguation methods in the literature focus on the task of re-assigning the edges of an already predefined set of ambiguous nodes, and the process
of identifying these nodes in a given non-attributed network, is usually overlooked. Thus, there exists very few approaches that tackle the latter case. In this section, we compare \acronym{} with three different competing approaches  that focus on the identification task, one existing method, and two heuristics.

\emph{Normalized-Cut (NC)}
The work of~\citep{saha2015name} comes close to ours, as their method also aims to identify ambiguous nodes in a given graph by
utilizing Markov Clustering to cluster an ego network of a vertex $u$ with the vertex itself removed. NC favors the grouping that gives small cross-edges between different clusters of $u$'s neighbors. The result is a score reflecting the quality of the clustering, using normalized-cut (\textbf{NC}).
$$NC=\sum^k_{i=1}\frac{W(C_i, \overline{C_i})}{W(C_i, C_i)+ W(C_i, \overline{C_i})}$$
with $W(C_i, C_i)$ as the sum of all the edges within cluster $C_i$, $W(C_i, \overline{C_i})$ the sum of the for all the edges between cluster $C_i$ and the rest of the network $\overline{C_i}$, and $k$ being the number of clusters in the graph.
% While~\cite{hermansson2013entity} also worked on identitying nodes based on topological features, their method performed worse in all the cases when compared to~\cite{saha2015name}. So we onnly chose the latter as a competing baseline.

\emph{Connected-Component Score (CC)}
We also include another baseline, Connected-Component Score (\textbf{CC}), relying on the same approach used in~\citep{saha2015name}, with a slight modification. Instead of computing the normalized cut score based on the clusters of the ego graph of a node, we account for the number of connected components of a node's ego graph, with the node itself removed.

\emph{Degree}
Finally, we use node degree as a baseline. As contracted nodes usually tend to have a higher degree, by inheriting more edges from combined nodes, degree is a sensible predictor for the node amibguity.

\ptitlenoskip{Evaluation Metric}
In the disambiguation literature, there has been no clear consensus on the use of a specific metric for the accuracy evaluation, but the most used ones vary between Macro-F1 and AUC.
We've performed evaluations for \acronym{} using the area under the ROC curve (AUC). A ROC curve is a $2$D depiction of a classifier performance, which could be reduced to a single scalar value, by calculating the value under the curve (AUC).
Essentially, the AUC computes the probability that our measure would rank a randomly chosen ambiguous node (positive example), higher than a randomly chosen non-ambiguous node (negative example).
Ideally, this probability value is $1$, which means our method has successfully identified ambiguous nodes $100\%$ of the time, and the baseline value is $0.5$, where the ambgiuous and non-ambiguous nodes are indistinguishable.
% It is a standard measure for prediction in Machine Learning.
This accuracy measure has been used in other works in this field, including~\citep{saha2015name}, which makes it easier to compare to their work.

\ptitlenoskip{Evaluation pipeline}\label{sec:pipeline}
% We evaluated the performance of the 4 different methods: \textbf{\acronym{}}, Normalized-Cut (\textbf{NC}), Connected-Components (\textbf{CC}), and \textbf{Degree}, already introcuded in~\ref{sec:baselinez}, on all the above datasets.%, except for DBLP.
We first perform network contraction on the original graph, by fixing the ratio of ambiguous nodes to $r$.
We then embed the network using CNE, and compute the disambiguity measure of \acronym{} Eq.~\ref{eq:obj_cne}, as well as the baseline measures for each node. Then the scores yield by the measures are compare to the ground truth (i.e., binary labels indicates whether a node is a contracted node.). This is done for 3 different values of $r\in\{0.001, 0.01, 0.1\}$.
We repeat the processes 10 more times using a different random seed to generate the contracted network and average the AUC scores.
For the embedding configurations, we set the parameters for CNE to $\sigma_1=1$, $\sigma_2=2$, with dimensionality limited to $d=8$.
\begin{figure}[H]
  \centering
      \includegraphics[width=0.5\textwidth]{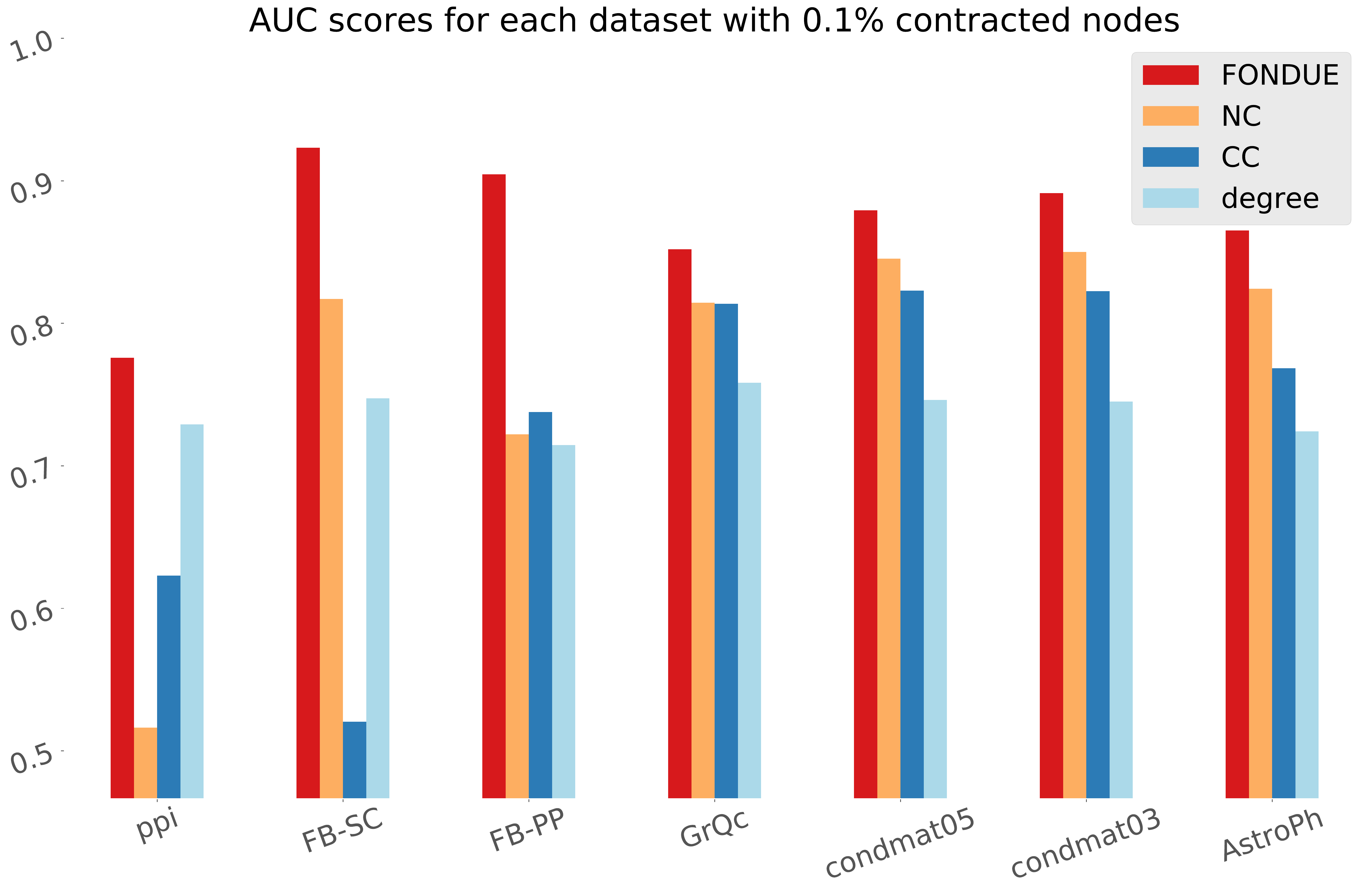}
    \includegraphics[width=0.5\textwidth]{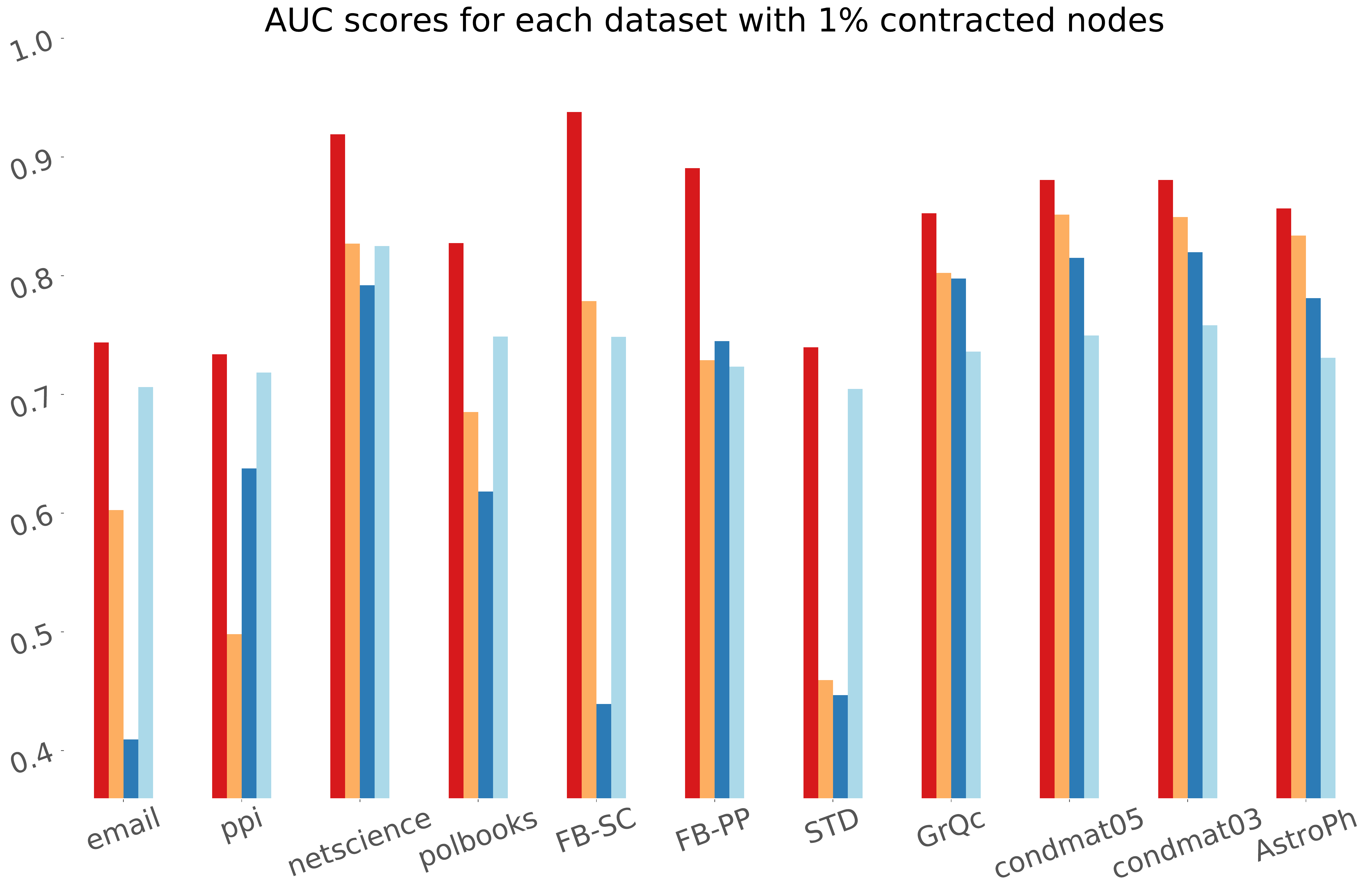}
    \includegraphics[width=0.5\textwidth]{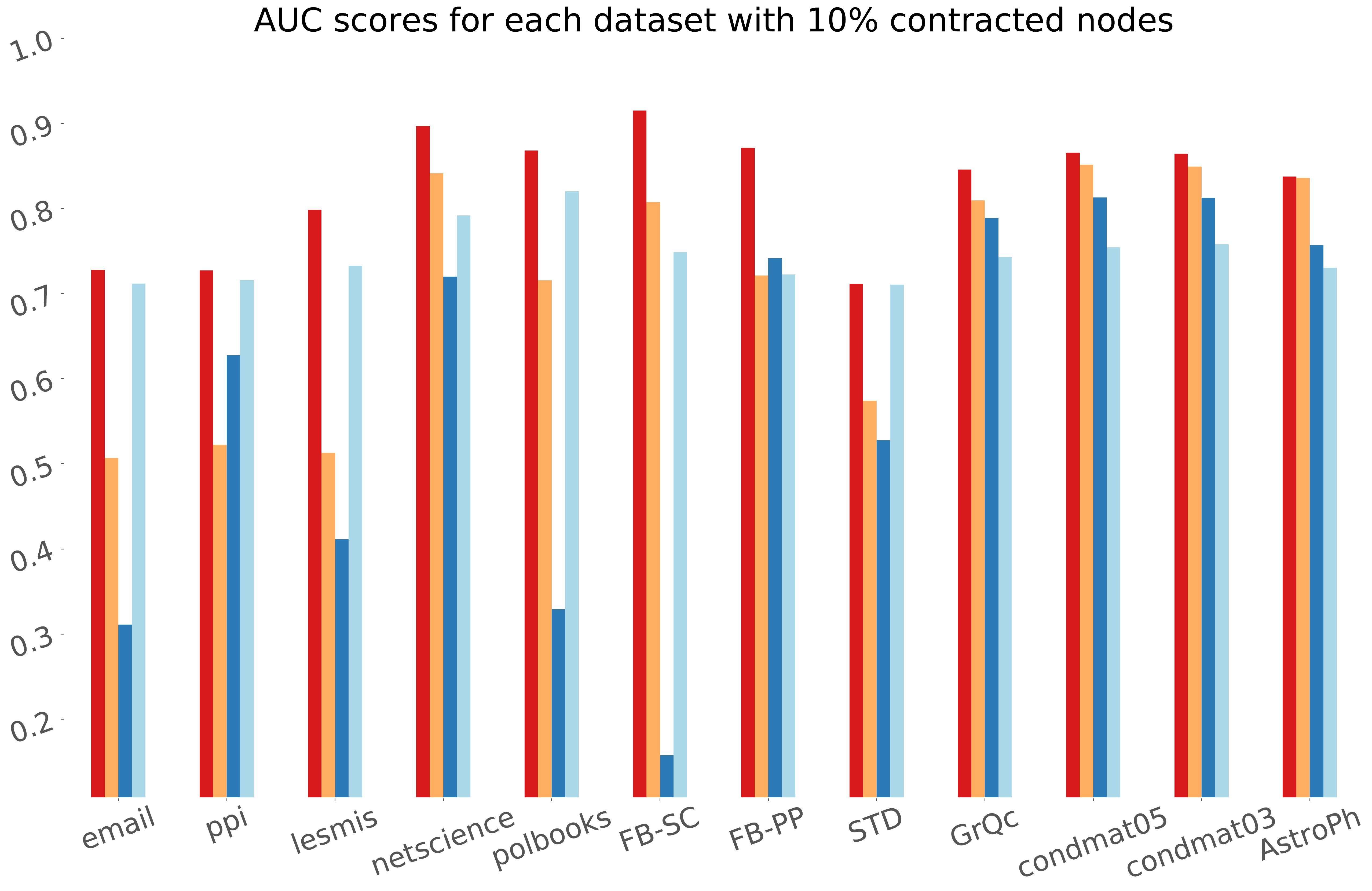}
      \caption{Bar plots for the AUC accuracy score for each dataset listed in Table~\ref{tab:quant}, for each of the 4 meaures, \acronym, Degree, NC, CC, for different percentage of contracted nodes, $0.1\%$, $1\%$,$10\%$ respectively.}
      \label{fig:quant}
\end{figure}

\ptitlenoskip{Results}\label{sec:results}
Results are illustrated in Figure~\ref{fig:quant} and shown in detail in Table~\ref{tab:quant}. \acronym{} outperforms the state-of-the-art method as well as non-trivial baselines in terms of AUC across all datasets. It is also more robust with various sizes of the networks, and the fraction of the ambiguous nodes in the graph. NC seems to struggle to identify ambiguous nodes for smaller networks (Table~\ref{tab:stat}). The results for node identification indeed address $\textbf{Q}_1$ and confirm the main contribution of this paper.
% For the DBLP dataset, as the ground truth was already present, \acronym{} scored $0.901$ on AUC compared to NC scoring $0.867$, which validates that our method is still outperforming NC for real world datasets, and confirm our results on the semi-synthetic datasets.
\begin{table}[]
\begin{adjustbox}{width=0.9\columnwidth,center}
  \begin{tabular}{@{}cllll@{}}
  \toprule
  \multicolumn{2}{c}{\textbf{\begin{tabular}[c]{@{}c@{}}\% of ambiguous \\ nodes\end{tabular}}} & \multicolumn{1}{c}{\textbf{dataset}} & \multicolumn{1}{c}{\textbf{FONDUE}} & \multicolumn{1}{c}{\textbf{MCL}} \\ \midrule
  \multicolumn{2}{c}{\multirow{8}{*}{10\%}}                                                      & fb-sc                                & 0.375                               & \textbf{0.776}                   \\
  \multicolumn{2}{c}{}                                                                          & email                                & 0.054                               & \textbf{0.239}                   \\
  \multicolumn{2}{c}{}                                                                          & student                              & \textbf{0.150}                      & 0.124                   \\
  \multicolumn{2}{c}{}                                                                          & lesmis                               & \textbf{0.402}                      & 0.304                            \\
  \multicolumn{2}{c}{}                                                                          & polbooks                             & 0.300                               & \textbf{0.338}                   \\
  \multicolumn{2}{c}{}                                                                          & ppi                                  & 0.025                               & \textbf{0.122}                   \\
  \multicolumn{2}{c}{}                                                                          & netscience                           & 0.503                               & \textbf{0.777}                   \\
  \multicolumn{2}{c}{}                                                                          & GrQc                                 & 0.398                               & \textbf{0.659}                   \\ \midrule
  \multicolumn{2}{c}{\multirow{6}{*}{1\%}}                                                     & fb-sc                                & 0.465                               & \textbf{0.847}                   \\
  \multicolumn{2}{c}{}                                                                          & email                                & 0.019                               & \textbf{0.281}                   \\
  \multicolumn{2}{c}{}                                                                          & student                              & 0.050                               & \textbf{0.183}                   \\
  \multicolumn{2}{c}{}                                                                          & ppi                                  & 0.016                               & \textbf{0.096}                   \\
  \multicolumn{2}{c}{}                                                                          & netscience                           & 0.609                               & \textbf{0.809}                   \\
  \multicolumn{2}{c}{}                                                                          & GrQc                                 & 0.638                               & \textbf{0.685}                   \\ \midrule
  \multicolumn{2}{c}{\multirow{4}{*}{0.1\%}}                                                    & fb-sc                                & 0.453                               & \textbf{0.917}                   \\
  \multicolumn{2}{c}{}                                                                          & ppi                                  & 0.030                               & \textbf{0.281}                   \\
  \multicolumn{2}{c}{}                                                                          & GrQc                                 & 0.722                               & \textbf{0.794}                   \\
  \multicolumn{2}{c}{}                                                                          & HepTh                                & 0.475                               & \textbf{0.569}                   \\ \bottomrule
  \end{tabular}
\end{adjustbox}
  \caption{Adjusted Rand Index score for \acronym{} and MCL}\label{tab:split}
\end{table}
\subsection{Quantitative Evaluation of Nodes Splitting}\label{sec:splitting}
Following the identification of the ambiguous nodes, how well does \acronym{} when it comes to
partitioning the set of edges into two separate ambiguous nodes. In this section, we focus on answering $\textbf{Q}_2$, node splitting.  Simply put, given an ambiguous node $v_i$, we refer to node splitting the process of replacing this particular node with two different nodes $\amb{v_i}, \ambb{v_i}$ and re-assigning the edges of $v_i$ such that $\Gamma(\amb{v_i}) \cup \Gamma(\ambb{v_i}) = \Gamma(v_i)$.

\ptitlenoskip{Baselines} For the node splitting task, the three baselines previously discussed in Sec~\ref{sec:baselinez} are not immediately applicable. However we adopt the Markov-Clustering (\textbf{MCL}) approach utilised in Normalized-Cut measure for splitting. Namely, a splitting is given by the MCL clustering on the ego network of an ambiguous node, with the node itself removed.

% \begin{figure}[H]
%   \centering
%       \includegraphics[width=0.5\textwidth]{{figs/random_start}.pdf}
%       \caption{Evaluate splitting quality using rand index score. When the number of random starts increases, the rand index score also increases. This indicates the splitting quality is correlated wit the embedding quality.}
%       \label{fig:random_start}
% \end{figure}

\begin{table*}
\caption{Execution time (in seconds) comparison table for the different datasets averaged over 10 different experiments}
\begin{adjustbox}{width=\textwidth,center}
  \begin{tabular}{@{}|l|ccc|ccc|ccc|@{}}
  \toprule
  \multicolumn{1}{|c|}{\textbf{\% of ambiguous nodes}} & \multicolumn{3}{c|}{\textbf{0.1\%}} & \multicolumn{3}{c|}{\textbf{1\%}} & \multicolumn{3}{c|}{\textbf{10\%}} \\ \midrule
  \multicolumn{1}{|c|}{\textbf{Methods}} & \multicolumn{1}{l}{\textbf{FONDUE}} & \multicolumn{1}{l}{\textbf{NC}} & \multicolumn{1}{l|}{\textbf{CC}} & \multicolumn{1}{l}{\textbf{FONDUE}} & \multicolumn{1}{l}{\textbf{NC}} & \multicolumn{1}{l|}{\textbf{CC}} & \multicolumn{1}{l}{\textbf{FONDUE}} & \multicolumn{1}{l}{\textbf{NC}} & \multicolumn{1}{l|}{\textbf{CC}} \\ \midrule
  \textbf{email} & - & - & - & 58.943 & 37.571 & 17.109 & 38.09 & 16.89 & 9.82 \\
  \textbf{ppi} & 159.435 & 153.905 & 55.046 & 158.276 & 141.338 & 43.183 & 135.10 & 113.66 & 45.77 \\
  \textbf{lesmis} & \textbf{-} & - & - & \textbf{-} & - & - & 0.29 & 0.21 & 0.04 \\
  \textbf{netscience} & \textbf{-} & - &  & 3.539 & 2.622 & 0.305 & 1.04 & 0.73 & 0.09 \\
  \textbf{polbooks} & \textbf{-} & - & - & 1.073 & 0.789 & 0.136 & 0.30 & 0.23 & 0.03 \\
  \textbf{FB-SC} & 818.305 & 231.161 & 282.695 & 826.875 & 212.404 & 287.885 & 422.67 & 525.24 & 173.79 \\
  \textbf{FB-PP} & 687.646 & 315.580 & 228.774 & 686.974 & 416.386 & 63.914 & 705.78 & 272.13 & 111.84 \\
  \textbf{STD} & \textbf{-} & - & - & 9.712 & 11.357 & 1.693 & 6.07 & 4.93 & 1.15 \\
  \textbf{GrQc} & 43.528 & 40.685 & 5.780 & 39.942 & 31.699 & 5.064 & 19.09 & 13.87 & 2.91 \\
  \textbf{condmat05} & 642.233 & 141.127 & 30.969 & 838.534 & 340.682 & 90.976 & 354.83 & 106.89 & 26.97 \\
  \textbf{condmat03} & 570.949 & 214.483 & 44.711 & 333.004 & 152.841 & 27.268 & 177.03 & 71.95 & 16.26 \\
  \textbf{AstroPh} & 402.846 & 348.269 & 84.526 & 337.125 & 162.863 & 96.724 & 189.76 & 81.04 & 53.26 \\ \bottomrule
  \end{tabular}
\end{adjustbox}
\label{tab:runtime}
\end{table*}
\ptitlenoskip{Evaluation Metric} Given a list of ambiguous nodes, we evaluate the splitting given by \acronym{} and MCL against the ground truth (node splitting according to the original network).
% As the goal is to identify the original ego network of the ambiguous nodes, by quantifying the quality of the clustering.
This is quantified by computing the Adjusted Rand Index (ARI) score between \acronym{} and the ground truth, as well as, between MCL and the ground truth.
ARI score is a similarity measure between two clusterings. %Here the clusterings are the ground truth as well as the splitting results obtained by \acronym{} and MCL.
ARI ranges between $-1$ and $1$, the higher the score the better the alignment between the two compared clusterings.
% 1 indicating a perfect match between the two sets of clusters, and $-1$ if  the index is less than the expected index.

\begin{figure}[H]
  \centering
      \includegraphics[width=0.5\textwidth]{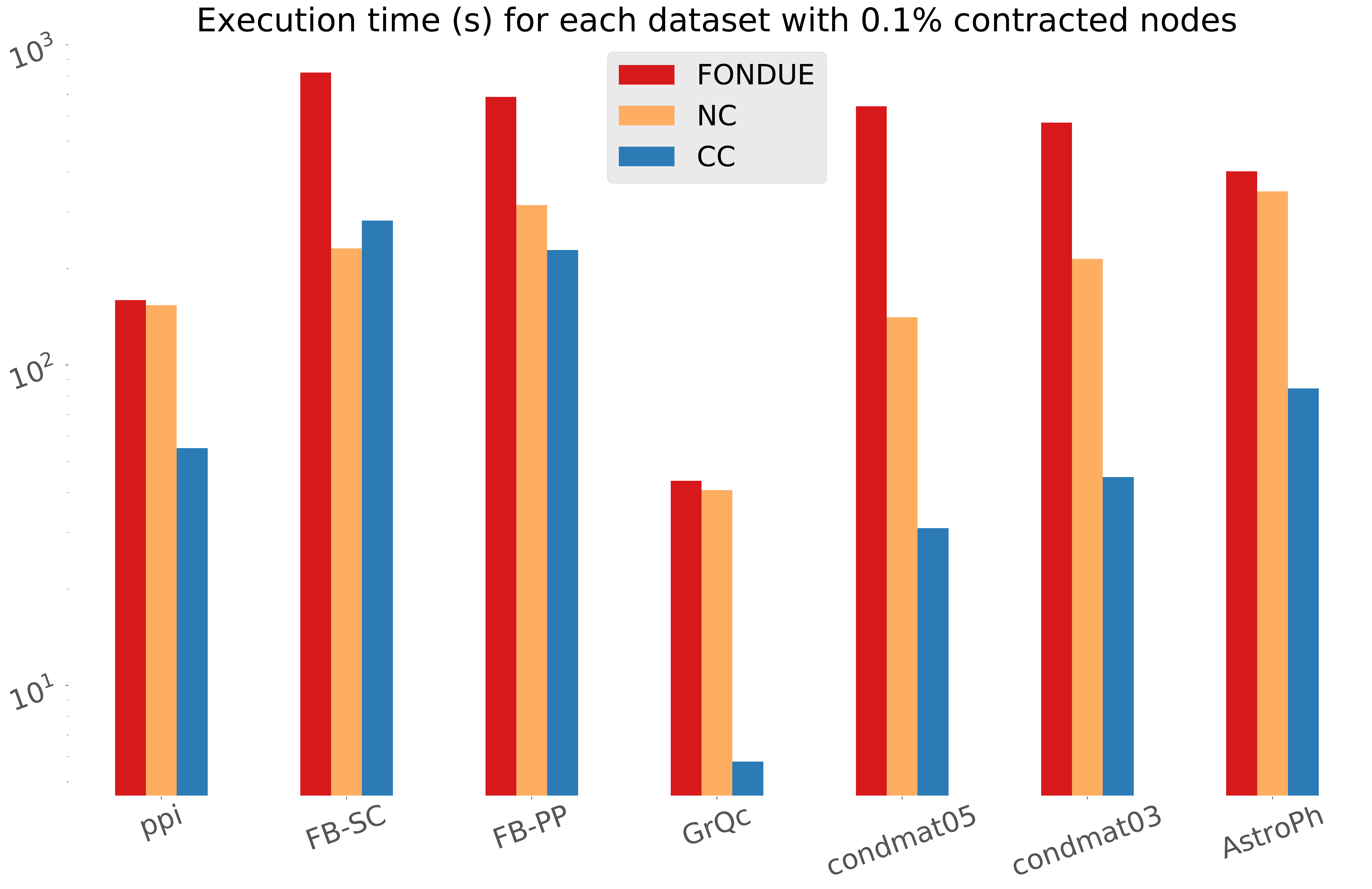}
    \includegraphics[width=0.5\textwidth]{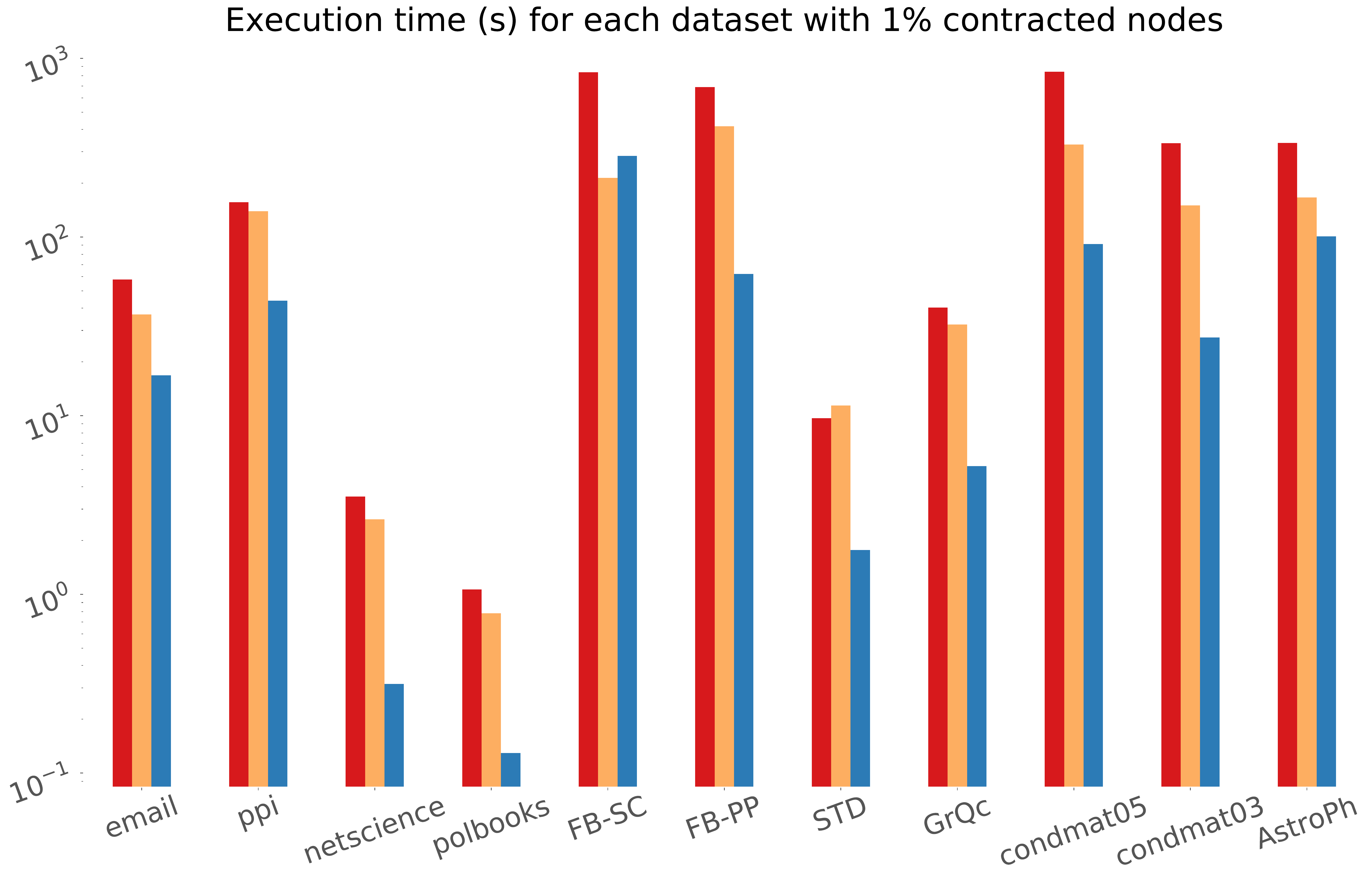}
    \includegraphics[width=0.5\textwidth]{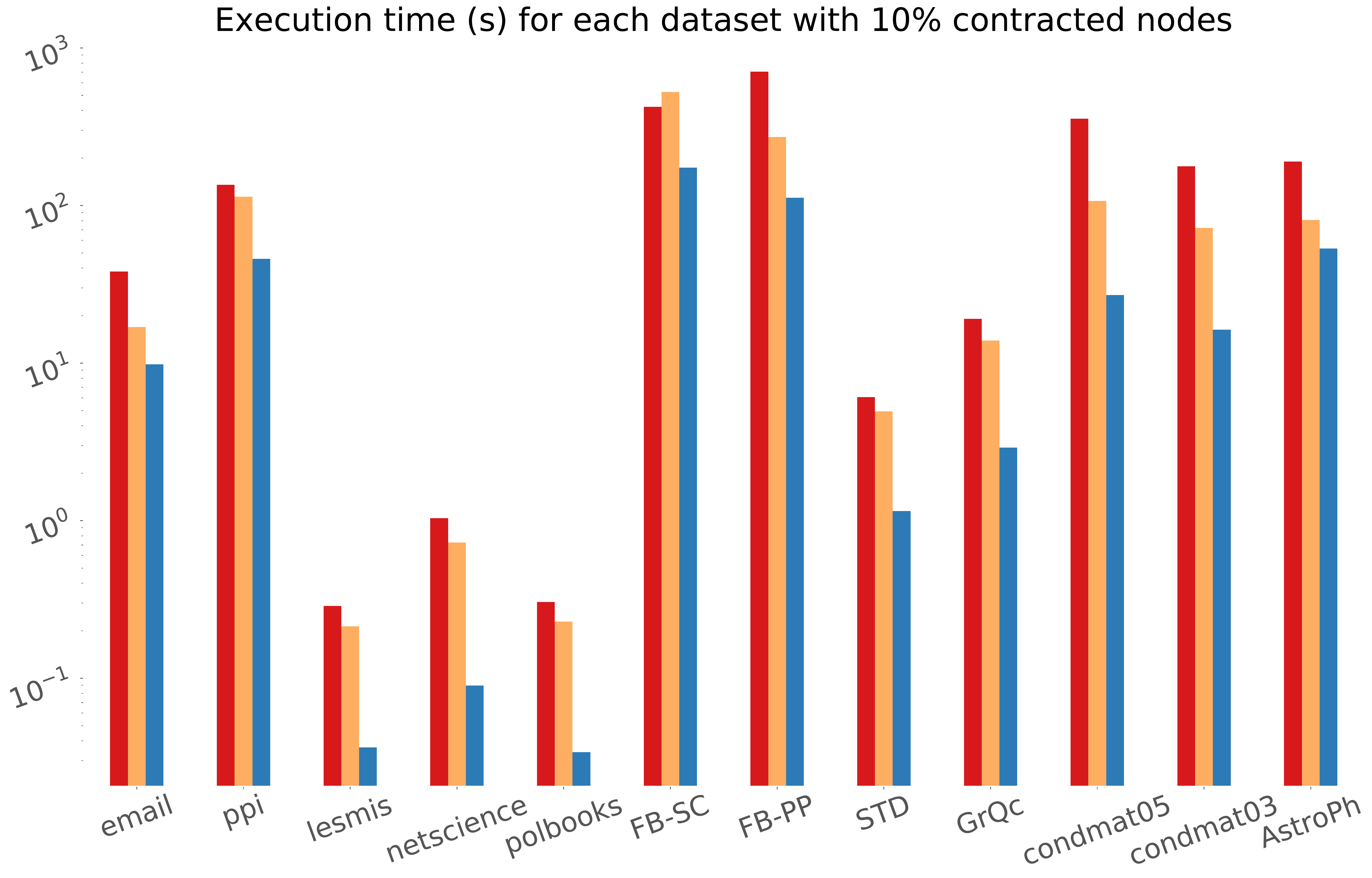}
      \caption{Bar plots for the runtime performance for each dataset listed in Table~\ref{tab:runtime}, for each of the 3 measures, \acronym, NC, CC for different percentage of contracted nodes, $0.1\%$, $1\%$,$10\%$ respectively.}
      \label{fig:runtime}
\end{figure}

\ptitlenoskip{Pipeline}
First we compute the ground truth. Then for each ambiguous node, we evaluate the quality (based on ARI) of the split from \acronym{} and MCL compared to the original partition. We repeat the experiments for three different contraction ratios $r\in\{0.001, 0.01, 0.1\}$ for each dataset. For each ration, the experiment is repeated three times with different random seeds.

\ptitlenoskip{Results}
Despite outperforming MCL in ambiguous node identification, \acronym{} seems to underperform compared to MCL on nearly every dataset Table~\ref{tab:split}.
The next step is to understand and diagnose the cause of the poor performance for node splitting.
Our initial suspicion veered towards either a poor optimization of the objective function, or that the Rayleigh Quotient~(\eqref{eq:quotient}) is not a good objective function. Upon further investigations, the latter seemed to be the cause of the poor performance in node splitting.

We verified our hypothesis using two approaches. Employing the MCL splitting results to evaluate the objective function, and computing the value of objective function for the ground truth compared to random sampling of the splitting on these nodes. Both experiments showed that our approximation method can always find a split with a higher Rayleigh Quotient objective value while the ground truth and MCL splitting scored lower.

We also suspect that low embedding quality might contribute to the underperfomance of \acronym{} in the splitting task, as one embedding often gets stuck in a local optimum.
So our further investigations went into choosing the best embedding that maximizes the CNE objective function out of 30 different random starts. The results showed that this can indeed, to a certain extent, improve, but not outperform MCL on the splitting task.
% And finally, we've performed incremental re-embedding of each node which also showed relative improvement on both tasks.

% To understand why, Further experiments are needed to investigate , but this lies beyond the scope of this paper and will be the main focus of our future work.

% \todo[inline]{MCL outperforms us, even tho we do better at identifying the ambiguous nodes than NC that relies on MCL. This is illustrated in table/figure whatever}
% Results are display in Table~\ref{tab:split}
% \begin{figure}
%   \centering
%       \includegraphics[width=0.5\textwidth]{{figs/dim_plot_0.1}.pdf}
%       \caption{Variation of the AUC with the dimension of the embeddings.}
%       \label{fig:cne}
% \end{figure}

\subsection{Parameter sensitivity}\label{sec:param-sens}
% \todo[inline]{How they change with different contraction ratio (Idenitication/Splitting)}
% Although \acronym{} is unsupervised and requires no tuning, the quality of the embeddings does affect its performance. %This is shown in Figure~\ref{fig:cne}.
% This is not a caveat of our method, but rather a curse of its modularity.\\
% We also evaluated the effects of the embedding dimension on the performance of \acronym{} \\
In this section, we study the robustness of our method against different network settings. Mainly how does the percentage of ambiguous nodes in a graph affect the node identification.
% and varying the number of random sampling for computing the objective function using the first approximation heuristic~\ref{sec:scale}.
In the previous experiments we've fixed the ratio of ambiguous nodes to $\{0.001, 0.01, 0.10\}$. we follow the same pipeline (generate, embed, evaluate for 10 different random seeds), for different ratios of ambiguous nodes.
% We study how the increase in the number of contracted nodes in the graph affect the accuracy performance each method.
As listed in Table~\ref{tab:quant} \acronym{} outperforms MCL and other baselines across nearly all networks with different contraction ratios.

\subsection{Execution time analysis}~\label{sec:runtime}
% \todo[inline]{They beat us (in a nice way)}
In Figure~\ref{fig:runtime}, we show the execution speed of \acronym{} and baselines in node identification and splitting. \acronym{} is slower than NC, but still comparable. Note that \acronym{} approximates \eqref{eq:quotient} by aggregating two different approximation heuristics (i.e., randomized, eigenvector thresholding Sec.~\ref{sec:scale}). The runtime results reflect the sum of the execution time of two heuristics. This is listed in details in Table~\ref{tab:runtime}. All the experiments have been conducted on a Intel $i7-7700$K CPU $4.20$GHz, running the Ubuntu 18.04 distribution of linux, with $32$GB of RAM.

%
% \subsection{Discussions}\label{sec:discuss}
%
% Despite its state-of-the-art performance in identifying ambiguous nodes (Sec.~\ref{sec:identifying}), \acronym{} node splitting ability is underwhelming compared to that of MCL (Sec.~\ref{sec:splitting}), contrary to our expectations.
% Nonetheless, we argue that's \acronym{}'s main prowness is to identify ambiguous nodes, which is highlight of the contribution of this paper, as its results are consistent accross different datasets and contraction ratio, rendering it a versatile tool for network ambiguity detection.
% %

\section{Conclusion}

In this paper we formalized the node disambiguation problem as an ill-posed inverse problem.
We presented \acronym{}, a novel method for tackling the node disambiguation problem,
aiming to tackle both the problem of identifying ambiguous nodes, and determining how to optimally split them.
\acronym{} exploits the empirical fact that naturally occurring networks can be embedded
well using state-of-the-art network embedding methods, 
such that the embedding quality of the network after node disambiguation can be used as an inductive bias.

Using an extensive experimental pipeline,
we empirically demonstrated that \acronym{} outperforms the state-of-the-art when it comes to the accuracy of identifying ambiguous nodes,
by a substantial margin and uniformly across a wide range of benchmark datasets of varying size, proportion of ambiguous nodes, and domain.
While the computational cost of \acronym{} is slightly higher than the best baseline method, the difference is moderate.

Somewhat surprisingly, the boost in ambiguous node identification accuracy was not observed for the node splitting task.
The reasons behind this (and potential approaches to remedy it) are subject of our ongoing research.
In the meantime, however, a combination of \acronym{} for node identification, and Markov clustering on the ego-networks of ambiguous nodes for node splitting,
is the most accurate approach to address the full node disambiguation problem.

\subsubsection*{Acknowledgements}

The research leading to these results has received funding from the European Research Council under the European Union's Seventh Framework Programme (FP7/2007-2013) / ERC Grant Agreement no. 615517, from the Flemish Government under the “Onderzoeksprogramma Artificiële Intelligentie (AI) Vlaanderen” programme, from the FWO (project no. G091017N, G0F9816N, 3G042220), and from the European Union's Horizon 2020 research and innovation programme and the FWO under the Marie Sklodowska-Curie Grant Agreement no. 665501.
This work was supported by the FWO (project numbers G091017N, G0F9816N, 3G042220).
% \subsubsection*{References}
% \clearpage
\bibliography{references}

\end{document}